\pdfoutput=1
 
\documentclass[10pt,conference,a4paper]{IEEEtran}
\usepackage{times,amsmath,epsfig}
\usepackage{booktabs}
\usepackage{color}
\usepackage{epstopdf}
\usepackage{multirow}
\usepackage{nicefrac}
\usepackage{soul}
\usepackage{subfigure}
\title{Neural Network Compression using Transform Coding and Clustering}
\author{%
{Thorsten Laude, Yannick Richter and J\"orn Ostermann} %  Removed for anonymous submission
{}
\vspace{1.6mm}\\
\fontsize{10}{10}\selectfont\itshape
Leibniz Universit\"at Hannover, Institut f\"ur Informationsverarbeitung, Germany\\ %  Removed for anonymous submission
\,\\ 
\\
\fontsize{9}{9}\selectfont\ttfamily\upshape
\vspace{1.2mm}\\
\fontsize{10}{10}\selectfont\rmfamily\itshape
\,\\ 
\\

\fontsize{9}{9}\selectfont\ttfamily\upshape
\,
}
\begin{document}
\maketitle

% INCLUDES COPYRIGHT NOTICE: one of three copyright notice should be included. Uncomment the appropriate line below, according to the authors affiliation:
\begin{figure}[b]
\parbox{\hsize}{\em
%information about the event:
%IEEE VCIP'14, Dec. 7 - Dec. 10, 2014, Valletta, Malta.

%copyright notice: one of three copyright notices below should be included. Uncomment the appropriate line, according to the authors affiliation:
%000-0-0000-0000-0/00/\$31.00 \ \copyright 2014 IEEE.
%U.S. Government work not protected by U.S. copyright.
%???-?-????-????-?/10/\$??.?? \copyright 2014 Crown.
}\end{figure}

\begin{abstract}
With the deployment of neural networks on mobile devices and the necessity of transmitting neural networks over limited or expensive channels, the file size of the trained model was identified as bottleneck.
In this paper, we propose a codec for the compression of neural networks which is based on transform coding for convolutional and dense layers and on clustering for biases and normalizations.
By using this codec, we achieve average compression factors between 7.9--9.3 while the accuracy of the compressed networks for image classification decreases only by 1\%--2\%, respectively. 
\\[1\baselineskip]
\end{abstract}

\section{Introduction}
\label{sec:intro}
The victory of the neural network \textit{AlexNet} \cite{Krizhevsky2012} in the ImageNet Large Scale Visual Recognition Challenge (ILSVRC) 2012 with a lead of more than 10 percentage points to the runner-up is considered by many as breakthrough for modern deep learning technologies.
Since then, deep neural networks spread to many scientific and industrial applications \cite{Sze2017} -- not only to image classification (e.g. \cite{He2016, Szegedy2015}), but also to speech recognition (e.g. Siri or Google Assistant), to augmented reality, etc. 
Often, the necessity of large amounts of training data, long training duration and the computational complexity of the inference operation are noted as bottlenecks in deep learning pipelines.

More recently, the memory footprint of saved neural networks was recognized as challenge for implementations in which neural networks are not executed on large-scale servers or in the cloud but on mobile devices (e.g. mobile phones, tablets) or on embedded devices (e.g. automotive applications).
In these use cases, the storage capacities are limited and/or the neural networks need to be transmitted to the devices over limited transmission channels (e.g. app updates).
Therefore, an efficient compression of neural networks is desirable.
This states true especially when it is considered that neural networks tend to become larger to cope with more and more advanced tasks.
General purpose compressors like Deflate (combination of Lempel-Ziv-Storer-Szymanski with Huffman coding) perform only poorly on neural networks as the networks consist of many slightly different floating-point weights.

The requirements for the compression of neural networks are:
a) high coding efficiency,
b) negligible impact on the desired output of the neural network (e.g. accuracy),
c) reasonable complexity, especially at the decoder (mobile device),
d) applicability to existing neural network models, i.e. no (iterative) retraining required.

In this paper, we propose a codec which addresses these requirements.
Namely, our main \textbf{contributions} are: 
\begin{itemize}
    \item a complete codec pipeline for the compression of neural networks,
    \item a transform coding method for the weights of convolutional and dense layers,
    \item a clustering-based compression method for biases and normalizations.
    %\item A bit depth-controlled quality setting
\end{itemize}

The remainder of this paper is organized as follows:
In Sec.~\ref{sec:related_work} we discuss related works from the literature and highlight the distinguishing features of our codec.
The details of our method are presented in Sec.~\ref{sec:method} and evaluated in Sec.~\ref{sec:evaluation}.
The paper ends with a conclusion in Sec.~\ref{sec:conclusion}.

\section{Related Work}
\label{sec:related_work}
Several related works were proposed in the literature.
These works mainly rely on techniques like quantization and pruning.

The \textit{tensorflow} framework provides a quantization method to convert the trained floating-point weights to 8 bit fixed-point weights. 
Similarly to this approach, we also utilize quantization.
In the evaluation (Sec. \ref{sec:evaluation}) we will demonstrate that considerable coding gains on top of those due to quantization can be achieved by our proposed methods.

Han \mbox{\textit{et al. }}proposed the \textit{Deep Compression} framework for the efficient compression of neural networks \cite{Han2015a}.
In addition to quantization, their method is based on an iterative pruning and retraining phase.
The weights and connections in the network which contribute the least to the output of the network are removed (pruning).
This by itself results in an unacceptable high decrease of the accuracy of the network.
Therefore, the network needs to be iteratively trained to reduce the accuracy drop.
In contrast to Deep Compression, we aim at transparent compression of existing network models without the necessity of retraining and without modifying the network architecture. 
Thereby, these approaches solve different problems.

Iandola \mbox{\textit{et al. }}propose a novel network architecture called \textit{SqueezeNet} which particularly aims at having as few weights in the network as possible \cite{Iandola2016}.
Thereby, their method is also not able to compress existing networks as it is our goal.
In the evaluation (Sec. \ref{sec:evaluation}), we will demonstrate that our method can still reduce the size of this already optimized SqueezeNet network by a factor of up to 7.4.

\section{Compression Method}

\label{sec:method}
\begin{figure}
    \centering
    \includegraphics[width=\columnwidth]{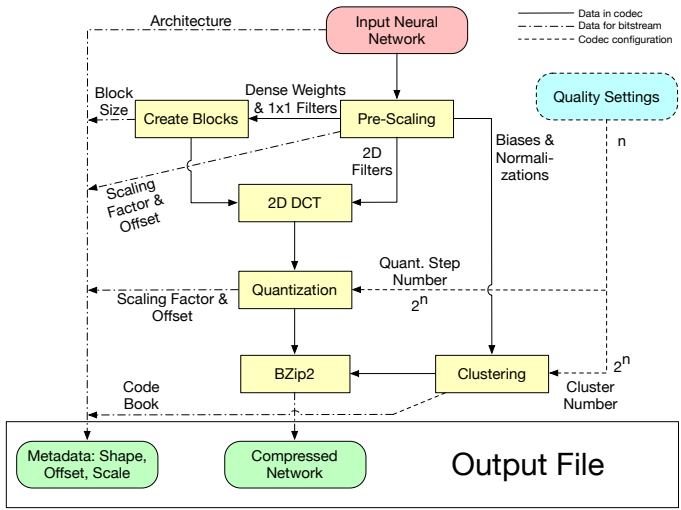}
    \caption{Pipeline of the proposed codec. Yellow blocks are applied per layer.}
    \label{fig:pipeline}
\end{figure}

\begin{figure}
    \centering
    \includegraphics[width=0.85\columnwidth]{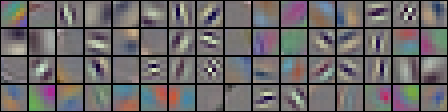}
    \caption{Exemplary filters from GoogLeNet}
    \vspace{-1.75\baselineskip}
    \label{fig:googlenet_filter}
\end{figure}

In this section, we describe our complete codec design with reference to the pipeline illustration in Fig.~\ref{fig:pipeline}.
The trained neural network model is the input of the codec.
It consists of one-dimensional and two-dimensional weights, biases, normalizations and the architecture itself (number of layers/filters, connections, etc.).
All layers of the network are processed individually.
This simplifies partly retroactive updates of individual layers without transmitting the complete network again. 
Depending on the available computing resources, an arbitrary number of layers can be processed in parallel.
Weights, biases and normalizations are pre-scaled to use the complete dynamic range of the number representation.
It is worth noting that we use the term pre-scaling here instead of normalization to avoid confusion with the term normalization which is already used for a type of network parameters.

Some exemplary filters from GoogLeNet are visualized in Fig.~\ref{fig:googlenet_filter}.
It is observable that the filters contain structural information not completely different from blocks in natural pictures.
Reasoned by this observation, the encoder base for convolutional filters consists of a two-dimensional discrete cosine transform (2D DCT) followed by a quantization step.
This combination is often referred to as transform coding.

For the DCT, the transformation block size is set accordingly to the size of the filter (e.g. a $7\times7$ DCT for a $7\times7$ filter).
Subsequent to the transformation, the coefficients are quantized.
Different from the transform coding in image compression (e.g. JPEG) where high frequency coefficients are quantized coarser than low frequencies to exploit properties of the human visual system, all coefficients are quantized equally in our codec.
This is due to the great importance of high frequencies like edges for the performance of the network.
The bit depth of the quantizer can be tuned according to the needs of the specific application.
Typical values are 5-6 bit/coefficient with only a small accuracy impact.

The weights of dense layers (also referred to as fully-connected layers) and of $1\times1$ convolutions (no spatial filtering but filtering over the depth of the previous layer, typically used in networks for depth reduction) are arranged block-wise prior to transform coding.
To achieve this, the one-dimensional weights are reshaped to the largest block size (up to a specified level of $8\times8$). 
Although these one-dimensional parameters do not directly have a spatial context, our research revealed that the transform coding still has a higher entropy-reducing impact than direct quantization.
Furthermore, it is worth noting that one-dimensional transform coding is not as efficient as  two-dimensional with the same number of values.

K-means clustering is used for the coding of the biases and normalizations.
It is worth noting that the outcome of this method is similar to using a Lloyd-Max quantizer.
The number of clusters is set analogously to the quantizer bit depth according to the quality settings.
Code books are generated for biases and normalizations. 
Thereby, the usage of the clustering algorithm is beneficial if less bits are needed for coding the quantizer indices and the code book itself than for coding the values directly.
The code book which is generated by the clustering algorithm depends on the initialization. 
For this reason, we run multiple iterations of the clustering with different initialization and select the code book which generates the smallest distortion.
It is our observation that ten iterations with 50 k-means steps each are sufficient (i.e. repeated simulations have the same results) and that more iterations do not considerably increase the performance. 
As the code book is signaled explicitly as part of the bit stream, the number of iterations has no influence on the decoding process.
The clustering approach has the advantage that the distortion is smaller than for uniform quantization. 
In consequence, the accuracy of the network is measured to be higher for a given number of quantizer steps.
However, the occurrence of code book indices is also more uniformly distributed.
Due to the higher entropy of this distribution, the compression factor is considerably smaller (see Sec.~\ref{sec:evaluation}).
In particular the Burrow-Wheeler transform and the move-to-front transform which are both invoked for entropy coding are put at a disadvantage by the uniform distribution.
We chose to use use the same number of quantizer steps for all parameters.
For this reason the clustering was chosen for those network parameters which are too sensible to the higher distortion caused by uniform quantization.

The processed data from the transform coding and from the clustering are entropy coded layer-wise using BZip2, serialized and written to the output file.
Padding to complete bytes is applied if necessary.
In addition to the weights, biases and normalizations, meta data is required for the decoding process and thus included in the output file as well.
It includes the architecture of the layers in the network, shapes and dimensions of the filters, details on the block arrangements, scaling factors from the pre-scaling, scaling factors and offsets from the quantizer, and the code books for the clustering.
This meta data stored in the file in addition to the network weights which is stored without structural information.
Thereby, the decoded network models can be loaded using the same APIs used for the original models.

\section{Evaluation}
\label{sec:evaluation}
In this section, we elaborate on the evaluation of our codec.
Without loss of generality, we select image classification as application since it is well-understood and the most common deep learning application.
Nevertheless, our methods are applicable to neural networks for other applications.
In total, we study the compression for four neural networks:
for the two state-of-the-art networks ResNet50 \cite{He2016} and GoogLeNet \cite{Szegedy2015}, for the famous AlexNet \cite{Krizhevsky2012} which has a special property as introduced below, and for SqueezeNet \cite{Iandola2016} whose architecture is already optimized to have as few weights as possible.

\begin{figure*}
	\centering
	\subfigure[GoogLeNet] {
		\begin{minipage}{0.4\linewidth}
			\centering
			\includegraphics[width=0.95\columnwidth]{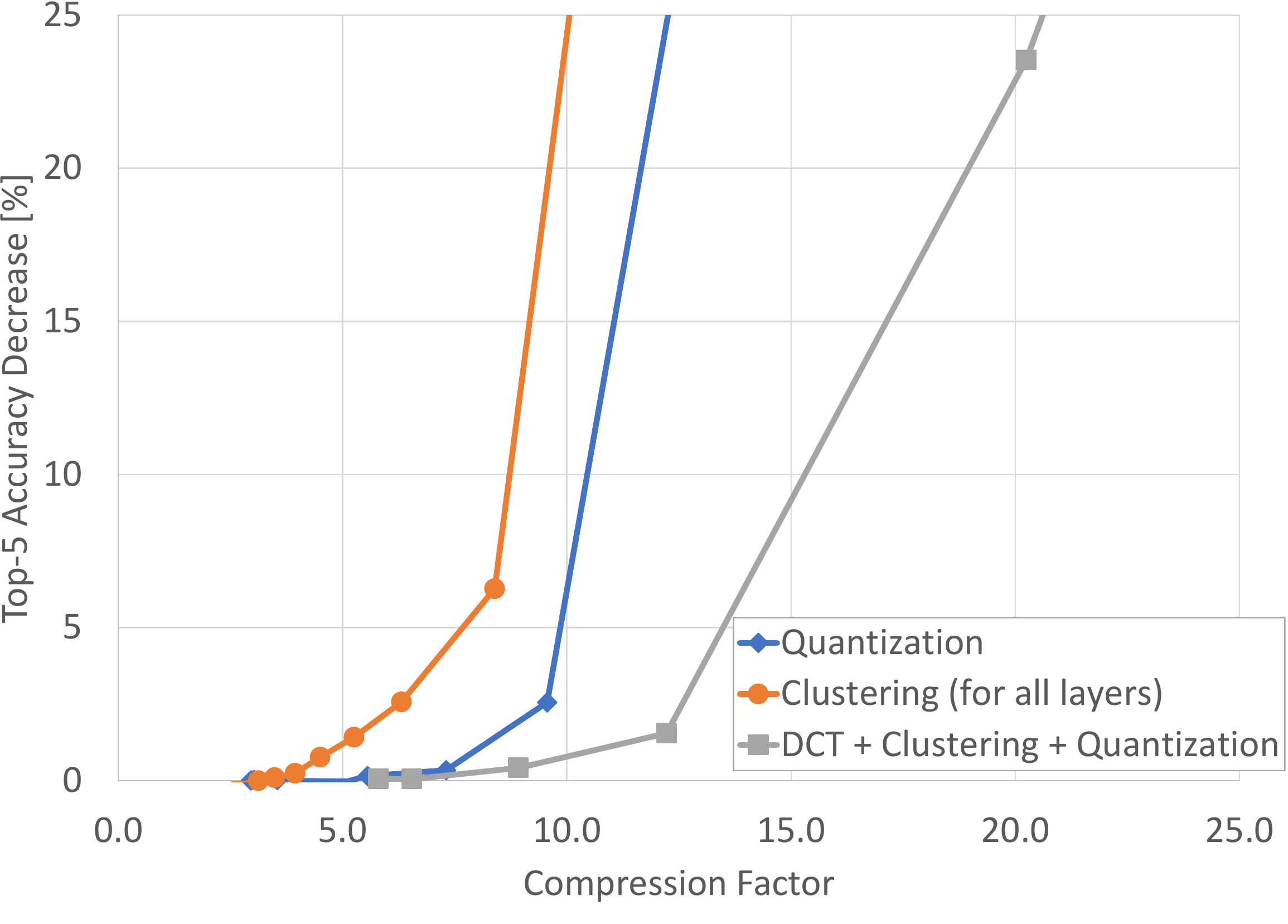}
		\end{minipage}
		\label{fig:rd_googlenet}
	}
	\subfigure[ResNet] {
		\begin{minipage}{0.4\linewidth}
			\centering
			\includegraphics[width=0.95\columnwidth]{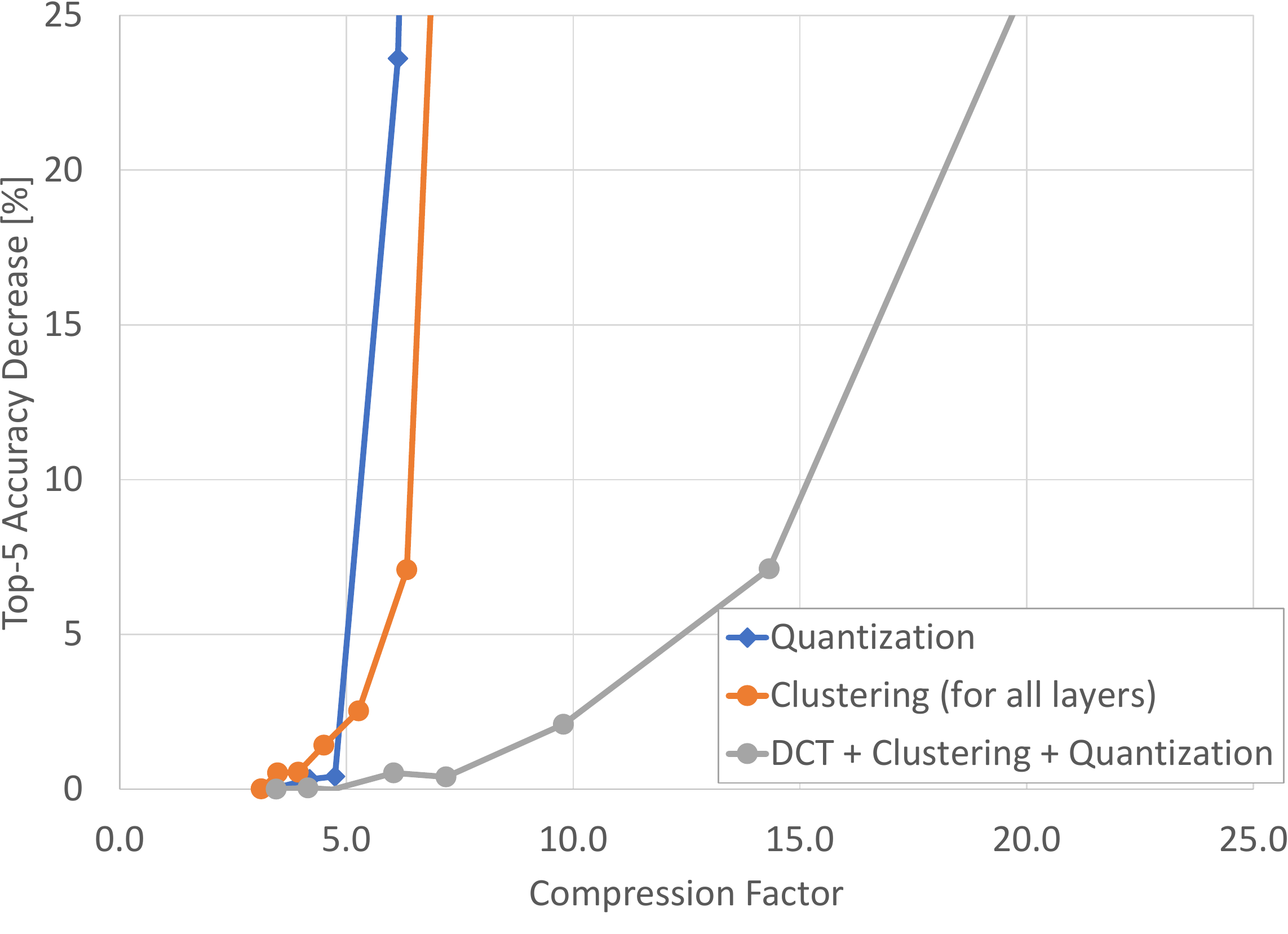}
		\end{minipage}
		\label{fig:rd_resnet}
	} \\
	\subfigure[AlexNet] {
		\begin{minipage}{0.4\linewidth}
			\centering
			\includegraphics[width=0.95\columnwidth]{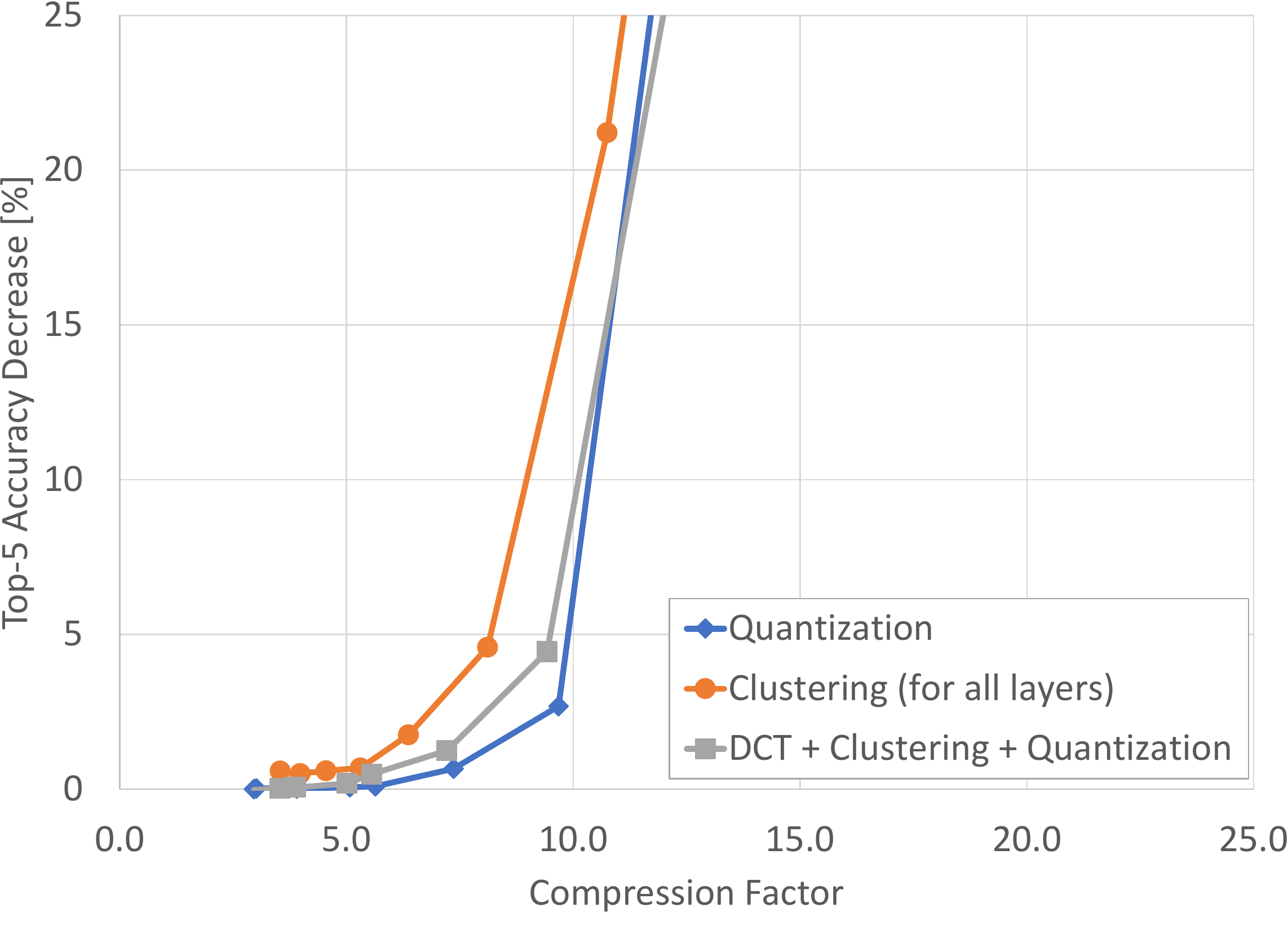}
		\end{minipage}
		\label{fig:rd_alexnet}
	}	
	\subfigure[SqueezeNet] {
		\begin{minipage}{0.4\linewidth}
			\centering
			\includegraphics[width=0.95\columnwidth]{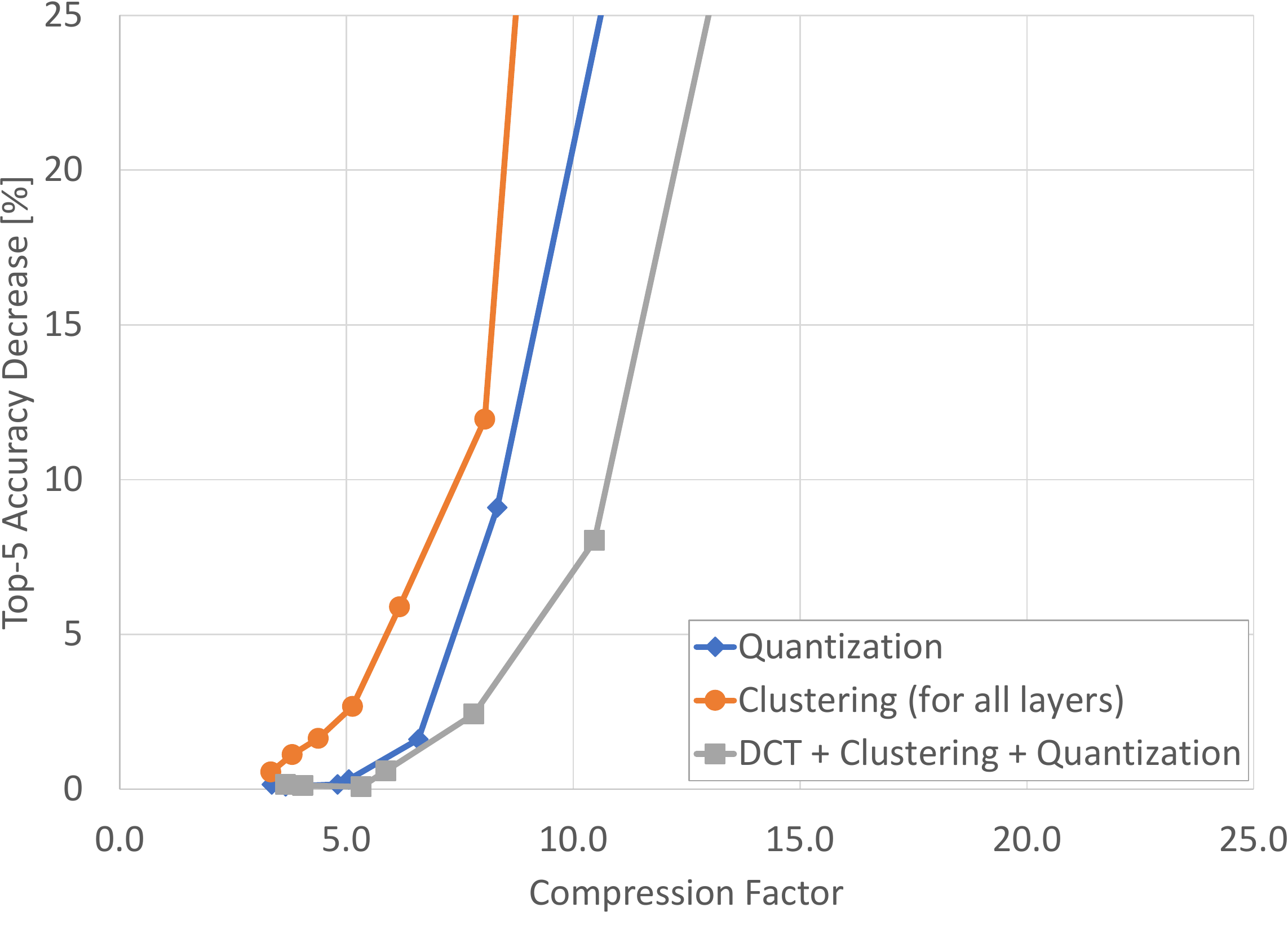}
		\end{minipage}
		\label{fig:rd_squeezenet}
	}	
	\caption{Top-5 accuracy decrease in percentage points as function of the compression factor ($\nicefrac{\textrm{uncompressed file size}}{\textrm{compressed file size}}$). Methods are better the more to the bottom-right they appear. BZip2 is incorporated for all methods.} 
	\vspace{-1.5\baselineskip}
	\label{fig:rd_plots}
\end{figure*}

% Table generated by Excel2LaTeX from sheet 'Interpolated Accuracy Drops'
\begin{table}
  \centering
  \caption{Compression factors for 1\% and 2\% decrease in accuracy}
    \begin{tabular}{c|l|c|c}
    \multirow{3}[1]{*}{Network} & \multicolumn{1}{c|}{\multirow{3}[1]{*}{Method}} & \multicolumn{2}{c}{Compression factor } \\
          &       & for 1\% & for 2\% \\
          &       & \multicolumn{2}{c}{accuracy decrease} \\
    \midrule[1pt]
    \multirow{3}[2]{*}{GoogLeNet} & Quantization & 8.0   & 9.0 \\
          & Clustering (for all layers) & 4.8   & 5.8 \\
          & DCT + Clustering + Quantization & \textbf{10.6} & \textbf{12.4} \\
    \midrule
    \multirow{3}[2]{*}{ResNet} & Quantization & 4.7   & 4.8 \\
          & Clustering (for all layers) & 4.2   & 4.9 \\
          & DCT + Clustering + Quantization & \textbf{8.1} & \textbf{9.7} \\
    \midrule
    \multirow{3}[2]{*}{AlexNet} & Quantization & \textbf{7.8} & \textbf{8.9} \\
          & Clustering (for all layers) & 5.6   & 6.6 \\
          & DCT + Clustering + Quantization & 6.7   & 7.7 \\
    \midrule
    \multirow{3}[2]{*}{SqueezeNet} & Quantization & 5.9   & 6.7 \\
          & Clustering (for all layers) & 3.7   & 4.6 \\
          & DCT + Clustering + Quantization & \textbf{6.3} & \textbf{7.4} \\
    \midrule[1pt]
    \multirow{3}[1]{*}{\textbf{Average}} & Quantization & 6.6   & 7.4 \\
          & Clustering (for all layers) & 4.6   & 5.5 \\
          & DCT + Clustering + Quantization & \textbf{7.9} & \textbf{9.3} \\
    \end{tabular}%
  \label{tab:compression_factors}%
\end{table}%

% Table generated by Excel2LaTeX from sheet 'Blatt1'
\begin{table}
  \centering
  \caption{Encoder and decoder run times}
    \begin{tabular}{l|cc}
    Method & \multicolumn{1}{l}{Encoding [s]} & \multicolumn{1}{l}{Decoding [s]} \\
    \midrule[1pt]
    Quantization & 26.2    & 6.5 \\
    Clustering (for all layers) & 383.0   & 7.2 \\
    DCT + Clustering + Quantization & 29.0  & 7.6 \\
    \end{tabular}%
    \vspace{-2\baselineskip}
  \label{tab:complexity}%
\end{table}%

Rate-distortion analysis is a typical procedure for the evaluation of compression algorithms. 
The performance of neural networks for image classification is usually measured using the Top-5 accuracy \cite{Sze2017}.
The higher the accuracy, the better the classification performance.
Therefore, we measure the distortion as decrease of the accuracy after compressing the networks.
Instead of using the bit rate or the absolute file size of the networks, we use the compression factor ($\nicefrac{\textrm{uncompressed file size}}{\textrm{compressed file size}}$) to assess the compression efficiency.
This facilitates the comparison of the results for different networks (with different file sizes) at a glance.

To generate RD curves, we sweep over the bit depth $n$ for the quantizer and the number of clusters ($=2^n$).
The bit depth is the same for all layers and not chosen adaptively.
The networks are encoded, decoded and then used for the image classification downstream pipeline.
As data, we use the ILSVRC-2012 validation set (50,000 images in 1,000 classes).
To study which algorithms from our overall pipeline contribute how much the the final result, we evaluate three subsets of our technology:
In the first subset, only quantization is applied to the network weights.
This is what is also used for the build-in compression of tensorflow.
In the second subset, we apply the clustering algorithm to all parameters of all layers.
In the third set, we use transform coding for the weights of convolutional and dense layers, and clustering for biases and normalizations.
The third set is the complete pipeline proposed in this paper.

The resulting RD curves are visualized in Fig.~\ref{fig:rd_plots}.
In this plot, it is desirable to be as far as possible on the right side (i.e. high compression factor) as well as on the bottom side (i.e. low accuracy decrease) as possible.
The findings for the two state-of-the-art networks (Fig.~\ref{fig:rd_googlenet} and~\ref{fig:rd_resnet}) are quite clear:
The results for the complete pipeline are superior to those of the subsets.
This indicates that all methods in the pipeline have a reason for existence and their coding gains are to some extend additive.
Compression factors of ten or higher are observed without a considerable decrease in accuracy.
AlexNet has the special property that it contains an extraordinary high number of weights and that more than 90\% of the weights are located in the first dense layer.
As suggested by Han \mbox{\textit{et al.}} \cite{Han2015a}, this disadvantage in the design of the network can only be fixed by pruning and retraining.
Hence, we observe in Fig.~\ref{fig:rd_alexnet} that transform coding and clustering do not bring any gain on-top of quantization.
Still, compression factors of 8--9 are observed since the entropy coding can achieve coding gains because the quantized weights contain lots of redundancy.
Interestingly, our methods enables compression factors of more then five without much decrease in accuracy even for SqueezeNet in Fig.~\ref{fig:rd_squeezenet}, a network whose architecture was already designed with the aim of having as few weights as possible (only 2\% of AlexNet).
This is an indication that our framework is also beneficial for networks with memory-optimized architecture. 

From the underlying data of Fig.~\ref{fig:rd_plots}, we calculate numerical values for the compression factor summarized in Tab.~\ref{tab:compression_factors}.
Accuracy decreases of 1\% or 2\% can be considered as acceptable for compressed networks in most applications since the compressed networks work almost as reliable as the original, uncompressed networks.
Therefore, we linearly interpolate compression factors for these two values.
In average, our codec (i.e. the complete pipeline) achieves compression factors of 7.9 and 9.3 for accuracy decreases of 1\% and 2\%, respectively.

We analyze the computational complexity of our algorithms by measuring the encoder and decoder run times. 
For this purpose, we execute the unoptimized Python code on an otherwise idle AMD Ryzen 2700x CPU for the GoogLeNet.
The run times for the three embodiments are summarized in Tab.~\ref{tab:complexity}.
As expected, applying only quantization is the fastest method with 26.2s for the encoder and 6.5s for the decoder.
The middle embodiment which involves clustering for all layers results in the highest encoding times by far (383s) which is due to the iterative clustering.
The decoding time does not increase considerably as the code book is not derived at the decoder but signaled as part of the bit stream.
For the final codec (transform coding for convolutional and dense weights, clustering for biases and normalizations), we measure 29s for the encoder and 7.6s for the decoder.
The encoding time is not increased much compared to sole quantization because the clustering is only applied to the biases and normalizations.
The decoder time increases slightly due to the inverse DCT in the decoding process.
Overall, both the encoder and decoder run times are acceptable, considering that the training of neural networks typically lasts days or weeks.
Merely for on-the-fly decoding faster run times would be desirable. 
This could easily be achieved by a performance-oriented implementation instead of using straight-forward Python implementations.

\section{Conclusion}
\label{sec:conclusion}
In this paper, we proposed a codec for the compression of neural networks.
The codec is based on transform coding for convolutional and dense layers, and clustering for biases and normalizations.
By using this codec, we achieve average compression factors between 7.9--9.3 while the accuracy of the compressed networks for image classification decreases only by 1\%--2\%, respectively.
%\vfill\eject

\bibliographystyle{IEEEtran}

\bibliography{IEEEabrv,References}

\end{document}